%% file: root.tex
\documentclass[letterpaper, 10 pt, conference]{ieeeconf}  % Comment this line out if you need a4paper

\IEEEoverridecommandlockouts                              % This command is only needed if 
\input{preamble}
\input{notation}

\title{\LARGE \bf Vision-aided Dynamic Quadrupedal Locomotion on Discrete Terrain using Motion Libraries}

\author{Ayush Agrawal$^{1}$, Shuxiao Chen$^{1}$, Akshara Rai$^{2}$ and Koushil Sreenath$^{1}$ %\textbf{author order TBD}% <-this % stops a space
\thanks{This work is supported in part by Meta AI through BAIR Open Research Commons and in part by National Science Foundation Grant CMMI-1944722.}% <-this % stops a space
\thanks{$^{1}$Authors are with the Department of Mechanical Engineering,
        University of California, Berkeley
        {\tt\small \{ayush.agrawal,  shuxiao.chen, koushils\}}@berkeley.edu}%
\thanks{$^{2}$Authors are with Meta AI,
        {\tt\small akshararai@fb.com}}%
}

\begin{document}
\setlength{\abovedisplayskip}{0pt}
\setlength{\belowdisplayskip}{0pt}%
\setlength{\textfloatsep}{1pt}	
\setlength{\abovecaptionskip}{0pt}

\maketitle
\thispagestyle{empty}
\pagestyle{empty}

%%%%%%%%%%%%%%%%%%%%%%%%%%%%%%%%%%%%%%%%%%%%%%%%%%%%%%%%%%%%%%%%%%%%%%%%%%%%%%%%
\begin{abstract}
In this paper, we present a framework rooted in control and planning that enables quadrupedal robots to traverse challenging terrains with discrete footholds using visual feedback. Navigating discrete terrain is challenging for quadrupeds because the motion of the robot can be aperiodic, highly dynamic, and blind for the hind legs of the robot. Additionally, the robot needs to reason over both the feasible footholds as well as the base velocity in order to speed up or slow down at different parts of the discrete terrain. To address these challenges, we build an offline library of periodic gaits which span two trotting steps, and switch between different motion primitives to achieve aperiodic motions of different step lengths on a quadrupedal robot. The motion library is used to provide targets to a geometric model predictive controller which outputs the contact forces at the stance feet. To incorporate visual feedback, we use terrain mapping tools and a forward facing depth camera to build a local height map of the terrain around the robot, and extract feasible foothold locations around both the front and hind legs of the robot. Our experiments show a small scale quadruped robot navigating multiple unknown, challenging and discrete terrains in the real world.
\end{abstract}
%%%%%%%%%%%%%%%%%%%%%%%%%%%%%%%%%%%%%%%%%%%%%%%%%%%%%%%%%%%%%%%%%%%%%%%%%%%%%%%%

\input{intro}
\input{hybrid_model}
\input{method}
\input{results}
\input{discussion}
\addtolength{\textheight}{-4cm}   % This command serves to balance the column lengths

%%%%%%%%%%%%%%%%%%%%%%%%%%%%%%%%%%%%%%%%%%%%%%%%%%%%%%%%%%%%%%%%%%%%%%%%%%%%%%%%
\bibliographystyle{ieeetr}
\bibliography{references}
\balance
\end{document}

%% file: preamble.tex
\usepackage{amsfonts}
\usepackage{amsmath}
\usepackage{amssymb}
\usepackage{graphicx}
\usepackage{xcolor}
\usepackage[english]{babel}
\def\inch#1{#1''}
\usepackage{caption}
\usepackage{subcaption}
\usepackage{booktabs}
\usepackage{balance}

\usepackage{mathtools}
\usepackage{hyperref}
\usepackage{scalerel,stackengine}
\stackMath
\newcommand\reallywidehat[1]{%
\savestack{\tmpbox}{\stretchto{%
  \scaleto{%
    \scalerel*[\widthof{\ensuremath{#1}}]{\kern-.6pt\bigwedge\kern-.6pt}%
    {\rule[-\textheight/2]{1ex}{\textheight}}%WIDTH-LIMITED BIG WEDGE
  }{\textheight}% 
}{0.5ex}}%
\stackon[1pt]{#1}{\tmpbox}%
}
\newtheorem{remark}{Remark}
\usepackage{svg}

%% file: notation.tex
\newcommand{\tvar}{t}
\newcommand{\position}{p}
\newcommand{\angularvel}{\omega}
\newcommand{\mass}{m}
\newcommand{\rotationmat}{R}

\newcommand{\lagrangian}{\mathcal{L}}

\newcommand{\var}{\delta}

\newcommand{\velocity}{\dot{\position}}
\newcommand{\inertia}{I}
\newcommand{\grav}{\mathbf{g}}
\newcommand{\rbstate}{\xi}
\newcommand{\rbinput}{F}

\newcommand{\rblinforce}{f}
\newcommand{\rbmoment}{\tau}

\newcommand{\timestep}{\Delta \tvar}
\newcommand{\work}{\mathcal{W}}
\newcommand{\deltarot}{\Delta \rotationmat}
\newcommand{\identity}{\mathbb{I}}
\newcommand{\zeros}{\mathbf{0}}
\newcommand{\quadinput}{\lambda_c}
\newcommand{\frictioncone}{\mathcal{K}_{\text{fric}}}

\DeclareMathOperator*{\argmin}{arg\,min}
\newcommand{\graspmap}{G_c}
\newcommand{\steplengthset}{\mathbb{L}}

%% file: intro.tex
\section{Introduction}
Legged robots have the unique capability to traverse across a wide variety of challenging and rough terrain, including terrains with gaps and discrete footholds. To navigate such terrain, a legged robot needs to precisely place its feet on feasible footholds, while maintaining its overall stability. This requires planning over multiple footsteps, and desired robot motion between the footsteps. For example, the robot might need to slow down and take a few steps on the same foothold, before speeding up and stepping over a large gap. Moreover, for unknown discrete terrain, the robot needs to make these decisions in real-time while navigating; stopping might make the robot unstable and gaps harder to cross. This results in a high-dimensional and complex optimization problem with a limited compute time budget. Discrete and uneven terrains also present an additional challenge of controlling the robot, as such terrains can result in the robot pitching, rolling and experiencing high angular velocities.

\subsection{Related Work}
Legged locomotion on discrete terrain, such as across stepping stones, is an active area of research with methods ranging from reduced-order models, to learning-based approaches. We summarize different research directions here: 

\textbf{{Reduced Order Models}}: In \cite{kajita2003biped}, the authors propose a reduced order cart-pole model to generate gaits for a bipedal robot to walk on randomly placed stepping stones. \cite{pratt2006velocity} presents a method to regulate the center-of-pressure to guide the robot leg onto a discrete foothold. More recently, in \cite{dai2021bipedal}, a reduced-order linear inverted pendulum model is presented to regulate the angular momentum about the stance foot at discrete impacts through the vertical center of mass velocity. A QP-based controller is then used to track outputs for 2D bipedal robots to walk on discrete terrain. 
\begin{figure}
    \centering
    \includegraphics[trim={2.5cm 1cm 0cm 4cm},clip,width=0.9\columnwidth, angle=0]{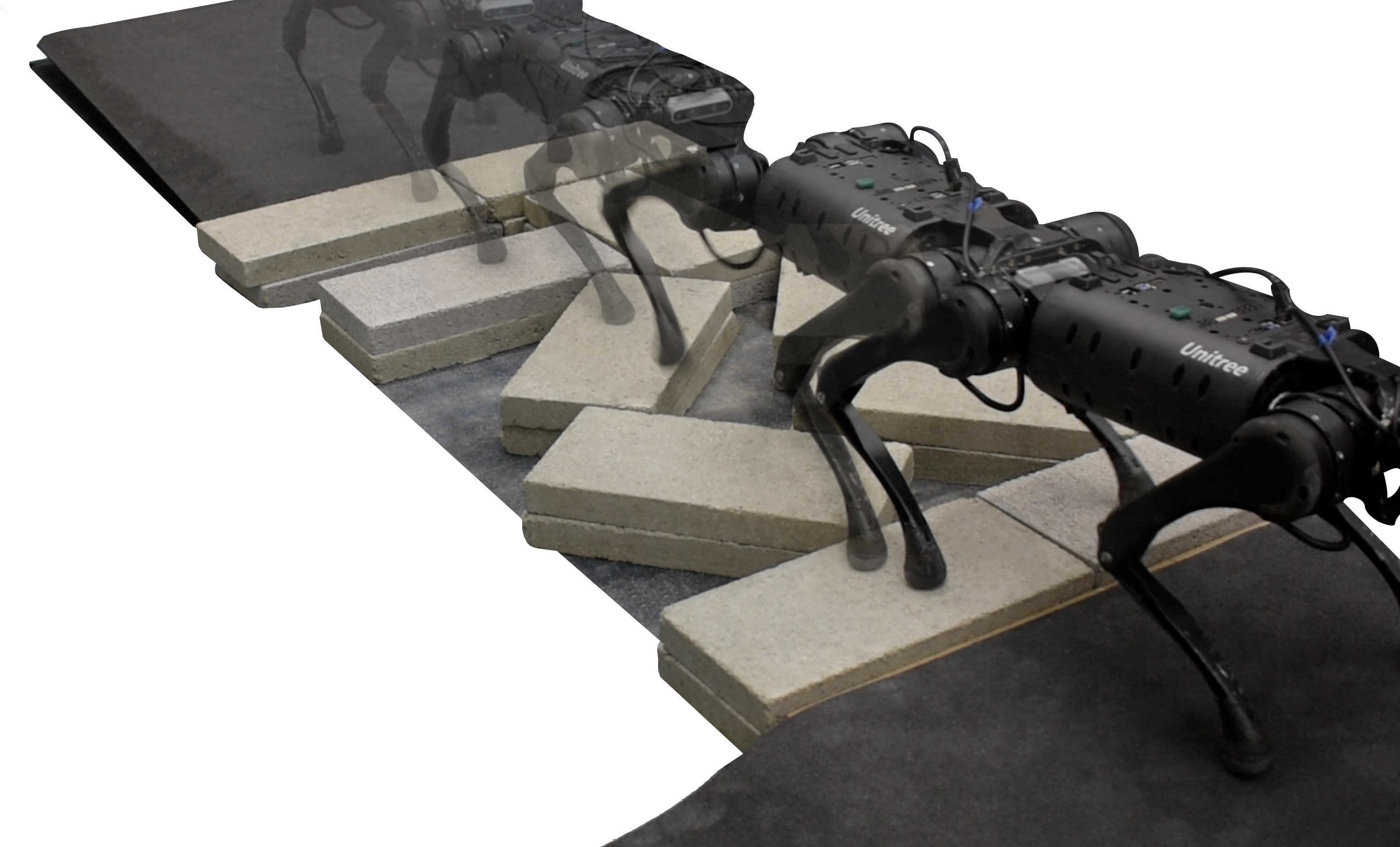}
    \caption{\small The A1 quadruped robot walking over a random discrete terrain using our proposed approach. Video of experiments can be found here: \href{https://youtu.be/3HAUvSsQYjs}{https://youtu.be/3HAUvSsQYjs}.}
    \label{fig:title_figure}
\end{figure}

\textbf{{Optimal Control}}: Optimization-based controllers such as Control Barrier Functions (CBFs) and Model Predictive Control (MPC) can enforce state and input constraints. In \cite{nguyen20163d,nguyen2015safety}, a CBF-based approach regulates the foot positions of a bipedal robot around a nominal periodic gait, to step on discrete footholds. This method is extended in \cite{nguyen2020dynamic} to use a library of walking gaits. The work in \cite{IJRR2018_ATRIAS_StochasticTerrainWalking,RSS2017_DiscreteTerrain_Walking,ECC2019_Running_on_SteppingStones} leverages the use of two-step periodic gaits, computed offline through trajectory optimization, to transition between gaits of different step lengths online. In \cite{grandia2020multi}, a multi-layered optimal control framework is presented that combines CBFs with MPC for precise foot placement over a planning horizon. Several other works \cite{mastalli2020crocoddyl,ishihara2019full,winkler2018gait} have also explored using trajectory optimization for dynamic legged locomotion.

\textbf{{Reinforcement Learning}}:
The work in \cite{tsounis2020deepgait} proposes to learn a high-level footstep planner, that takes in the local height-map of the terrain as input, and outputs a sequence of desired footstep locations. A low-level joint controller is then learned to track these footsteps. In \cite{xie2021glide}, the authors propose learning the desired accelerations for a centroidal model of a quadruped and use a heuristic approach to plan for footsteps on discrete terrain. The work in \cite{xie2020allsteps} proposes a curriculum with varying levels of difficulty to learn a policy for various bipedal robots to walk across stepping-stones. Several methods such as in \cite{siravuru2017deep,yu2021visual,margolis2021learning} and \cite{villarreal2020mpc} have also explored combining learning based approaches, particularly for vision-based footstep planning along with model-based low-level joint control. 

\subsection{Approach and Primary Contributions}
In this work, we study the problem of dynamic locomotion for quadrupedal robots across discrete terrain, using visual feedback. Our primary contributions in this work are threefold. First, we extend our prior work for bipedal robots in \cite{IJRR2018_ATRIAS_StochasticTerrainWalking,RSS2017_DiscreteTerrain_Walking,ECC2019_Running_on_SteppingStones}, for solving footstep selection problem by building a library of \emph{two-step periodic} gaits, to quadrupedal robot locomotion. By pre-computing an offline library of two-step periodic gaits, parametrized by the step lengths in the first and second steps, transition between different step lengths can be achieved online by switching between the different motion primitives. Moreover, with trajectory optimization tools such as \cite{hereid2019rapid}, an offline library with several hundred gaits can be computed within tens of minutes. Unlike bipedal robots, however, additional kinematic constraints exist between the front and hind limbs of the quadruped. To overcome this, we propose to create a motion library of two-step-periodic \emph{trotting} gaits comprising of four stance phases (equivalent to four bipedal steps).

Next, to stabilize these gaits, we propose a novel coordinate-free MPC that considers the evolution of the orientation of the robot on the $SO(3)$ manifold, as opposed to an Euler angle representation. Different from existing methods, we develop a discrete-time model using principles from Geometric Variational Integrators \cite{siravuru2018reaction} that preserves the inherent geometric structure of the $SO(3)$ manifold.

Finally, using terrain mapping frameworks \cite{Fankhauser2014RobotCentricElevationMapping,Fankhauser2018ProbabilisticTerrainMapping} with a forward facing depth camera, we incorporate visual feedback and experimentally validate our proposed approach on a \texttt{Unitree A1} quadrupedal robot to navigate across multiple unknown, challenging discrete terrains.

%% file: hybrid_model.tex
\section{Hybrid Model of Trotting}

In this section, we introduce the necessary background and notations for the robot dynamics model considered in our approach. Our formulation of the dynamics is derived from prior work on hybrid dynamics, as in \cite{RSS2017_DiscreteTerrain_Walking}.

\textbf{Configuration Variables}: The \texttt{Unitree A1} is a $10$kg quadruped with 3 motors in each leg, with a total of 12 actuated joints and 6 underactuated base degrees of freedom (DoF). The configuration of the robot is represented by $q = \left[p^T, \Theta^T, q_{FR}^T, q_{FL}^T, q_{RR}^T, q_{RL}^T \right]^T\in \mathcal{Q} \subset \mathbb{R}^{18}$, where $p \in \mathbb{R}^3$ denotes the Cartesian position of the robot, $\Theta \in \mathbb{R}^3$ denotes the $ZYX$ Euler angle representation of the orientation of the body, and $q_i \in \mathbb{R}^3, ~ i \in \lbrace FR, FL, RR, RL \rbrace$  denotes the actuated joints of the front/rear right/left legs. The actuated joints include hip abduction, hip and knee pitch DoFs. 

\textbf{Continuous Dynamics}: 
The dynamics model for each phase of a trotting gait can be obtained through the method of Lagrange and represented by the \emph{Manipulator equations}:
\begin{align}
    D(q) \ddot{q} + C(q, \dot{q})\dot{q} + G(q) &= B\tau + J_c^T\lambda_c, \label{eq:cont_dyn}\\
    J_c \ddot{q} + \dot{J}_c \dot{q} &\equiv 0, \nonumber 
\end{align}
\noindent
where $D$ is the inertia matrix, $C$ the Coriolis terms, $G$ gravitational terms, $B$ a selection matrix. $J_c$ denotes the contact Jacobian, $\lambda_c$ denotes the contact forces at the feet, and $\tau \in \mathbb{R}^{12}$ denotes the motor torques. The dimension of $J_c$ and $\lambda_c$ depend on the phase of gait, and number of legs in contact with the ground.

\textbf{Impact Dynamics}: The collision of the feet with the ground is modelled as an instantaneous rigid impact and the post-impact velocities $\dot{q}^+$ is obtained by solving the linear system of equations
\begin{align}
    \begin{bmatrix}
    D(q) & -J_c^T(q)\\
    J_c(q) & \mathbf{0}
    \end{bmatrix} \cdot
    \begin{bmatrix}
    \dot{q}^+\\
    \lambda_c
    \end{bmatrix} = \begin{bmatrix}
    D(q)\dot{q}^-\\
    \mathbf{0}
    \end{bmatrix}. \label{eq:impact_dyn}
\end{align}

\textbf{Hybrid Model:} We model each trotting step with two alternating phases of double-support (DS), where the diagonal feet are in contact, and quadruple-support (QS), where all four feet are in contact, as illustrated in Fig. \ref{fig:trotting-step}. Combining \eqref{eq:cont_dyn} and \eqref{eq:impact_dyn}, we obtain a hybrid dynamical model for trotting as,
\begin{align}
	\Sigma_{ds} &: \begin{cases}
		\dot{x}  &= f_{ds}(x) + g_{ds}(x)\tau, ~x \notin \mathcal{S}_{ds\rightarrow qs}\\
		x^+ &= \Delta_{ds \rightarrow qs}\left(x^-\right),~x \in \mathcal{S}_{ds\rightarrow qs}
	\end{cases}\nonumber\\
	\Sigma_{qs} &: \begin{cases}
		\dot{x}  &= f_{qs}(x) + g_{qs}(x)\tau, ~ (x, \tau) \notin \mathcal{S}_{qs\rightarrow ds}\\
		x^+ &= \Delta_{qs \rightarrow ds}\left(x^- \right), ~ (x, \tau) \in \mathcal{S}_{qs\rightarrow ds}
	\end{cases},\label{eq:hybrid_dyn}
\end{align}
\noindent
where $x := [q^T, \dot{q}^T]^T$ is the state of the robot, $f_{ds}(x)$, $g_{ds}(x)$ and $f_{qs}(x)$, $g_{qs}(x)$ denote the vector-fields in the DS and QS domains respectively, and are obtained from \eqref{eq:cont_dyn}. The switching surface $\mathcal{S}_{ds\rightarrow qs} := \lbrace x ~|~ p_{sw}^z(x) = 0, \dot{p}_{sw}^z(x) < 0 \rbrace$ is defined to be the set of states where the vertical component of the swing foot position is zero and the vertical swing foot velocity is less than zero. $\mathcal{S}_{qs\rightarrow ds}:= \lbrace (x, \tau) ~|~ \lambda_c^z(x, \tau) = 0\rbrace$ corresponds to the set of states and control inputs where the vertical ground reaction force at the stance feet $\lambda_c^z(x, \tau) \equiv 0$ (when the stance foot lifts off from the ground). The reset map $\Delta_{ds\rightarrow qs}$ is obtained from impact dynamics \eqref{eq:impact_dyn} and $\Delta_{qs\rightarrow ds} = \mathbb{I}$ is the identity operator. 
\vspace{-0.29cm}

%% file: method.tex
\section{Approach}
We now present our proposed approach of trajectory optimization and model-based low-level robot control using geometric MPC. We begin by building a motion library consisting of gaits parametrized by step length and optimized to minimize the total energy over a step, subject to dynamics and periodicity constraints. The low-level controller takes optimized CoM trajectories and footstep locations from the gait library to generate desired joint torques that are applied on the robot. Lastly, we describe the localization and terrain mapping framework we use in real-world experiments.
\subsection{Trajectory Optimization}
\begin{figure}
    \centering
    \includegraphics[width=0.45\textwidth]{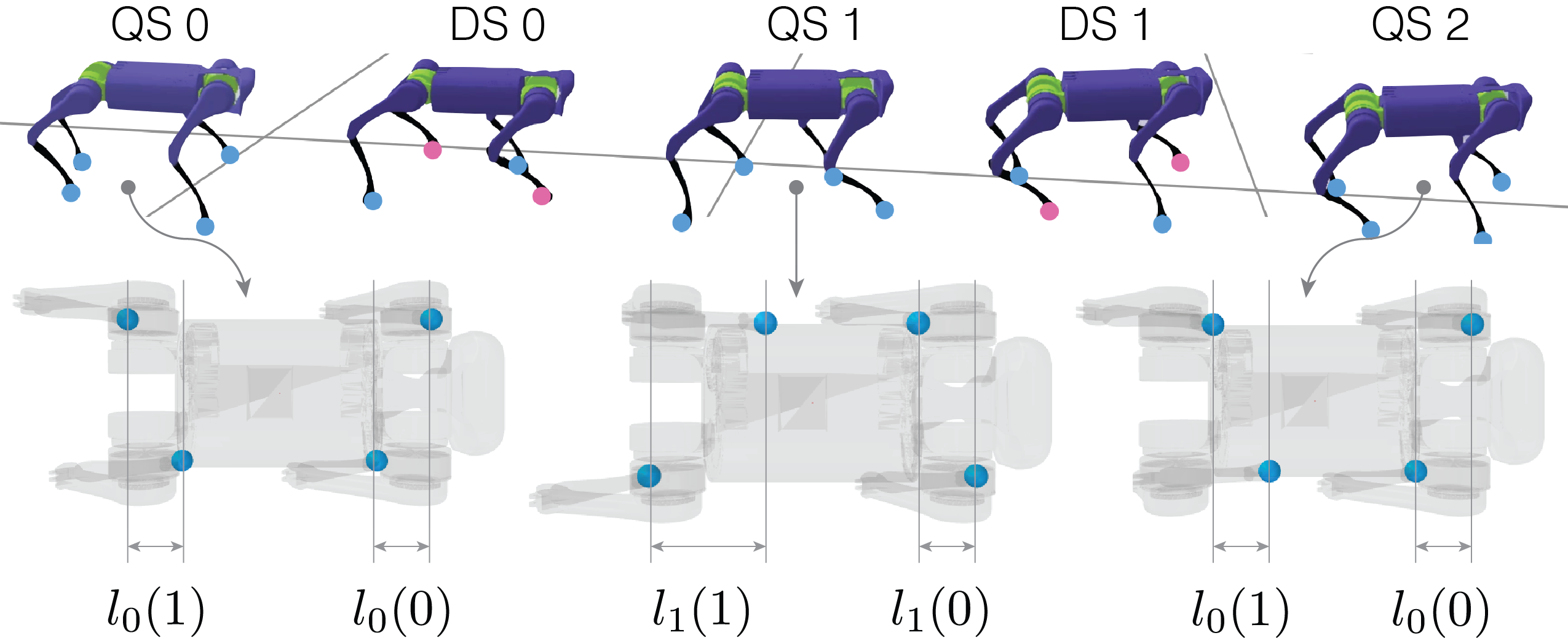}
    \caption{(Top) A trotting gait consists of two DS and QS domains as indicated by the figures marked from QS 0 to DS 1. For a `one-step' periodic gait, the state at the beginning of the next step (QS 2) must coincide with the initial state of the previous step (QS 0). (Bottom) A `one-step' periodic trotting gait is overly restrictive to capture all possible transitions between $l_0$ and $l_1$. When $l_0$ and $l_1$ are chosen independently, `one-step' periodic solutions for a trotting gait do not exist (the configuration of the robot in QS 2 does not coincide with the configuration in QS 0). To obtain `one-step' periodic trotting gaits, $l_0$ and $l_1$ are constrained by $l_0(0) + l_1(1) = l_0(1) + l_1(0)$. A `two-step' periodic trotting gait used in this paper consists of four DS and four QS phases and provides sufficient flexibility to choose $l_0$ and $l_1$ independently.} \label{fig:trotting-step}
\end{figure}

In this section, we present a method to generate a motion library of trotting gaits that achieve foot-placements of different step lengths. In particular, we obtain gaits that are `two-step' periodic, such that the post-impact states of the robot after two trotting steps return to the initial states at the start of the first step. The gaits are parametrized by the step-lengths $l_0, l_1 \in \mathbb{R}^2$ as indicated in Fig. \ref{fig:trotting-step}. The step lengths $l_0$ and $l_1$ each represent a pair of distances between the left and right pairs of feet. The goal of trajectory optimization is to find gait parameters $\gamma \left(l_0, l_1\right)$ for various step-length pairs to construct a library of gaits denoted by $\mathcal{G} := \lbrace \gamma\left(l_0, l_1\right)~|~ \left(l_0, l_1 \right)\in \steplengthset\times \steplengthset\rbrace$, where $\steplengthset \coloneqq L\times L$ is a predefined set of step length pairs. Specifically, we choose  $L = \lbrace -0.2, -0.1, 0.0, 0.1, 0.2 \rbrace$m, with a total of $5^4$ gaits in the library. $\gamma\left(l_0, l_1 \right)$ comprises of the trajectory parameters for the base linear and angular velocities, and body height and orientation, which serve as reference states for the MPC as detailed in Section \ref{sec:mpc}.

The trajectory optimization problem is solved using Direct Collocation which involves discretizing each phase in time by a specified number of nodes $N$ \cite{hereid20163d},  with the objective of minimizing energy over the entire trajectory, subject to dynamics and additional constraints $c_i(x_i(t), \tau_i(t))$, 
	\begin{align}\label{eq:traj_opt}
		(x^*(\cdot), \tau^*(\cdot)) &= \argmin_{x(t), \tau(t)}\quad  \Sigma_i \int_{0}^{T} ||\tau(t)||^2_2~ dt \\
		\mathrm{st.}~ &x(t) = \int_0^T f_i(x(t)) + g_i(x(t))\tau(t) dt, \nonumber \\
		& c_i(x(t), \tau(t)) \leq 0, \quad 0 \leq t \leq T,  ~ \forall i \in \mathcal{I}. \nonumber
	\end{align}
Here, $\mathcal{I}$ denotes the set of all discrete phases, $c_i(x(t), \tau(t))$ encodes physical constraints such as state and input limits, friction constraints as well as periodicity and step length constraints. 
The desired gait parameters $\gamma\left(l_0, l_1\right)$ can then be extracted from the optimal state trajectories $x^*(\cdot)$. We use the open-source toolbox \texttt{C-FROST} \cite{hereid2019rapid} to model and solve the above optimization problem. We refer the reader to \cite{RSS2017_DiscreteTerrain_Walking} for specific details on the trajectory optimization formulation. 

\begin{remark}
The generation of periodic gaits for quadrupeds with varying step lengths poses additional challenges and constraints compared to bipedal robots in \cite{RSS2017_DiscreteTerrain_Walking}. These challenges arise from kinematic constraints between the left and right limbs. In particular, to independently choose the step-length pairs $l_0$ and $l_1$, and to also induce periodicity constraints, we consider `two-step' periodic trotting gaits which comprise of four QS and four DS phases (as opposed to `one-step' periodic gaits). To visualize the requirement of a `two-step' periodic gait, we first consider a `one-step' periodic gait. When $l_0$ and $l_1$ are chosen independently, the net displacements of the left and right pairs of feet during a step is not necessarily equal. As a result, the four feet of the robot can move closer together or further apart during a step, resulting in a gait that is \emph{not} periodic. This is illustrated in Fig. \ref{fig:trotting-step}, where the chosen step-lengths $l_0(0)=l_0(1)=l_1(0)=0.1m$ and $l_1(1)=0.2m$ result in the four feet moving closer together at the end of a step (the configuration in QS 2 does not coincide with QS 0). Additional constraints on $l_0$ and $l_1$ must be placed to obtain `one-step' periodic gaits. In particular, the net displacements of the left and right pairs of feet during a step must be equal. This is captured by the constraint  $l_0(0) + l_1(1) = l_0(1) + l_1(0)$. A `two-step' periodic gait, on the other hand, consists of two additional DS and QS phases. By appropriately choosing the step-lengths in these phases, the net displacements of the left and right pairs of feet in two steps can be made equal, while still being able to choose $l_0$ and $l_1$ independently.  
\end{remark}
\vspace{-0.15cm}
\subsection{Footstep Planning and Gait Selection}
\label{sec:footstep}
Once we have created the gait library, we can extract desired gait variables by querying motions that satisfy the environment foothold constraints and that start from the current state of the robot.

\textbf{Footstep Planning}: To chose a desired foothold location, we first query the gait library  to obtain a nominal foothold location based on the current configuration and center-of-mass velocity of the robot as well as a nominal desired center-of-mass velocity. Similar to \cite{xie2021glide}, we then chose a desired step-length that is closest to the nominal foothold location and on the feasible terrain. 

\textbf{Gait Selection}:  Given the current state of the robot and the feasible footstep map, we extract a gait from the library based on the current step-length $l_0$ and the desired step-length $l_1$ through bi-linear interpolation of the gait library \cite{RSS2017_DiscreteTerrain_Walking}. This returns the desired states for a reduced-order rigid-body model considered in the MPC controller.  This update allows us to re-target the desired CoM velocities to be consistent with the desired step-lengths. 
\vspace{-0.1cm}
\input{geometric_mpc}
\vspace{-0.1cm}
\subsection{Swing leg control}
For the swing-leg control, we implement an output PD controller to follow a desired foot trajectory,
\begin{equation}
    \tau_{sw} = J_{sw}^{T}\left(-K_p^{sw}(p^{sw} - p^{sw}_d) -K_d^{sw}(\dot{p}^{sw} - \dot{p}^{sw}_d)\right).
\end{equation}

The desired foot trajectories are parametrized by B\'ezier polynomials such that the initial desired position is located at the true foot position at the start of a swing phase, and the final position based on the desired step-length, obtained through a foot-step planner (Section \ref{sec:footstep}).  
\vspace{-0.1cm}
\subsection{Localization and Mapping}
We use a forward facing depth camera to perceive the terrain, which makes it challenging to pick feasible footsteps for hind limbs. This requires building a local map of the robot by fusing a history of depth images that the robot sees, as well as the estimate of its own inertial pose in order to build a local map of the terrain around the robot. We fuse two libraries to achieve this:

\textbf{Localization}: We implement contact-aided invariant EKF from \cite{hartley2020contact} to localize the robot in the world. The binary contact information, required by the EKF, is obtained through contact force sensors located at the feet.

\textbf{Mapping}: We utilize the probabilistic robot-centric mapping framework developed in \cite{Fankhauser2018ProbabilisticTerrainMapping,Fankhauser2014RobotCentricElevationMapping} to obtain a height-map of the terrain. Localization estimates from the EKF and depth images from the robot camera are fused by the mapper to build a local map around the robot. We distinguish between stepable and un-stepable terrain based on the height and normal direction and add a 5cm threshold at the edges between these regions to account for inaccuracies in the foot placement controller and state estimation. The local map based on previously observed depth images is used for picking footholds for the hind limbs, eliminating the issue of lack of perception towards the back of the robot. In the future, this localization and mapping framework can be replaced by learning-based approaches, which can automatically build a history of feasible footholds.
\vspace{-0.1cm}

%% file: geometric_mpc.tex
\subsection{Geometric Model Predictive Control for Stance Legs}\label{sec:mpc}

We now present our Geometric MPC framework, which outputs the contact forces of the stance legs, with the objective to stabilize the robot's CoM trajectory and body orientation. MPC is a widely used method to control quadrupedal robots, but requires a linearization of CoM dynamics to simplify the underlying optimization for efficient real-time computation. A common approach to linearizing the CoM dynamics involves small angle approximation of the body roll and pitch and a \emph{Jacobian} linearization of the orientation dynamics \cite{di2018dynamic}. 
However, the small-angle approximation restricts the domains in which the model is valid, especially on uneven terrain where the robot might experience high angular velocities and pitch due to disturbances. Additionally, since the dynamics of the robot body evolve on the $SE(3)$ manifold, singularity issues arise in the Jacobian linearization process. Euler discretization of the continuous-time orientation dynamics also results in the loss of the underlying geometric structure of the $SO(3)$ manifold and, as a result, the discrete-time dynamics are not energy preserving \cite{siravuru2018reaction}. 

This has led to research in geometric variation-based optimal control approaches \cite{chignoli2020variational, wu2015variation, hong2020real} that linearize the quadruped dynamics using \textit{rotation matrices} instead of Euler angles. The resulting linearization is coordinate free, and does not suffer from singularities. However, \cite{chignoli2020variational} does not consider discrete time dynamics of the linearized system required for MPC, and \cite{hong2020real} use forward Euler to discretize the orientation dynamics. Euler discretization of the orientation dynamics, however, results in the loss of important mechanical properties like energy and momentum conservation, and the discrete-time dynamics may not evolve in the $SO(3)$ manifold \cite{siravuru2018reaction}. 

We present Geometric Variational MPC (GVMPC) which applies a variation-based linearization \cite{wu2015variation} to a reduced-order model of the quadruped, while ensuring that the discretized system is energy conserving. Similar to prior works, we model the quadruped as a single rigid body actuated by linear forces and moments about its CoM. 

\textbf{Discretization:} We begin by formulating a discrete-time model of the rigid-body dynamics as required by the MPC. Inspired by \cite{siravuru2018reaction} we consider the system Lagrangian discretized using the Trapezoidal rule with a time step of $\timestep \coloneqq t_{k+1} - t_k$:
\vspace{-0.1cm}
\begin{align}
    \lagrangian_k \approx \int_{\tvar_k}^{\tvar_{k+1}}\lagrangian d\tvar = \lagrangian \timestep,
\end{align}
\noindent
where $\lagrangian$ is the Lagrangian in continuous-time. To obtain the discrete-time dynamics of the system, we equate the action-sum to zero, $\Sigma_{k=0}^{N-1}\var\lagrangian_k + \var \work_k = 0$. $\var \work_k$ is the infinitesimal work done by the force $\rblinforce_k$ and moment $\rbmoment_k$, 
  \begin{align}
    \var \work_k := \timestep \left(\rblinforce_k \cdot \var \position_k + \rbmoment_k \cdot\var \eta_k\right), \label{eq:action_sum}
 \end{align}
 where $\var \position_k$ is an infinitesimal displacement and $\var \eta_k \in \mathbb{R}^3$ can be interpreted as an infinitesimal change in orientation. 
   \begin{figure}%[t]
    \centering
    \begin{subfigure}[b]{0.23\textwidth}
        \centering
        \includegraphics[width=0.95\textwidth]{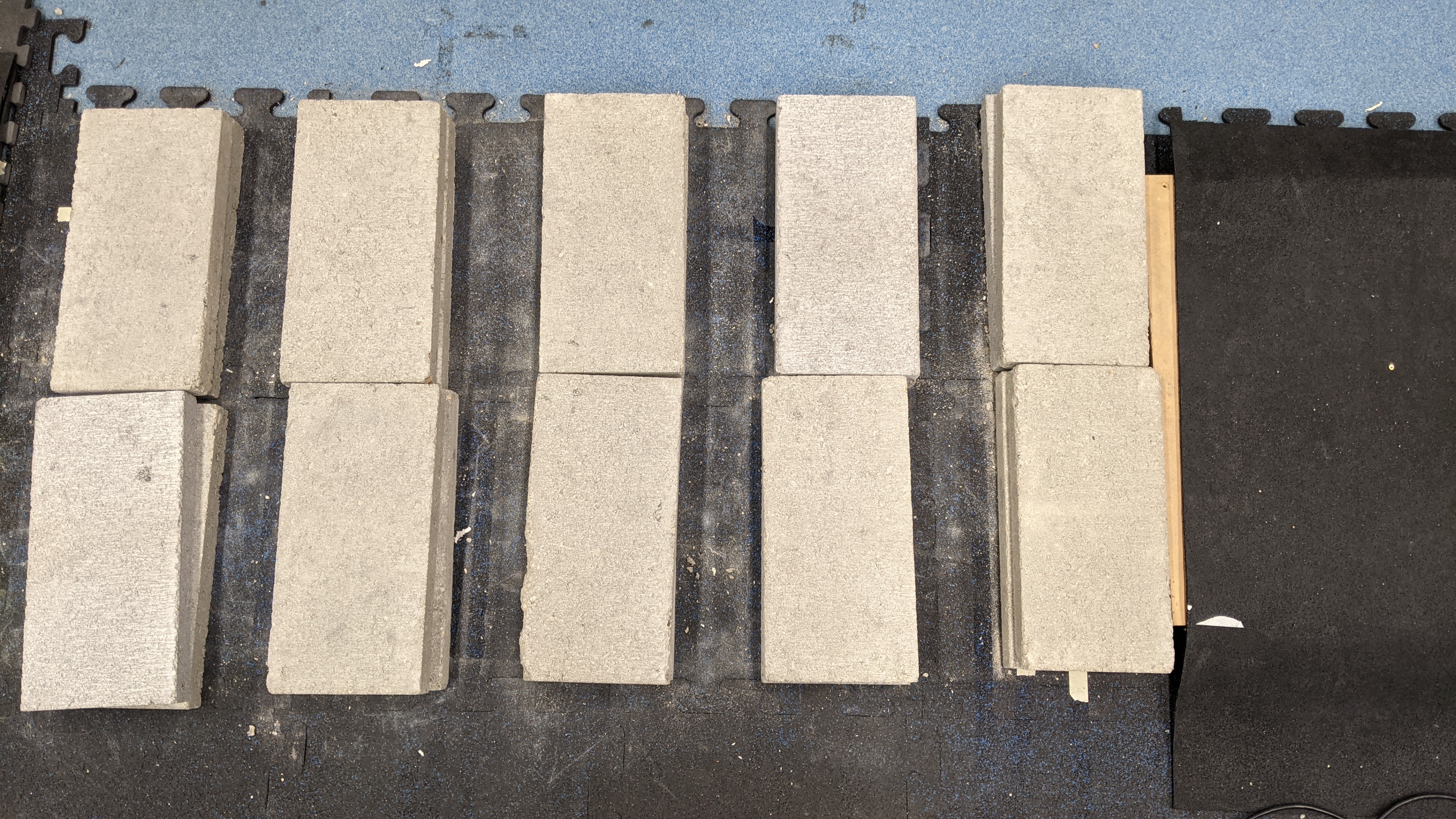}
        \caption{Aligned terrain}
        \label{fig:aligned}
    \end{subfigure}
    \begin{subfigure}[b]{0.23\textwidth}
        \centering
        \includegraphics[width=0.95\textwidth]{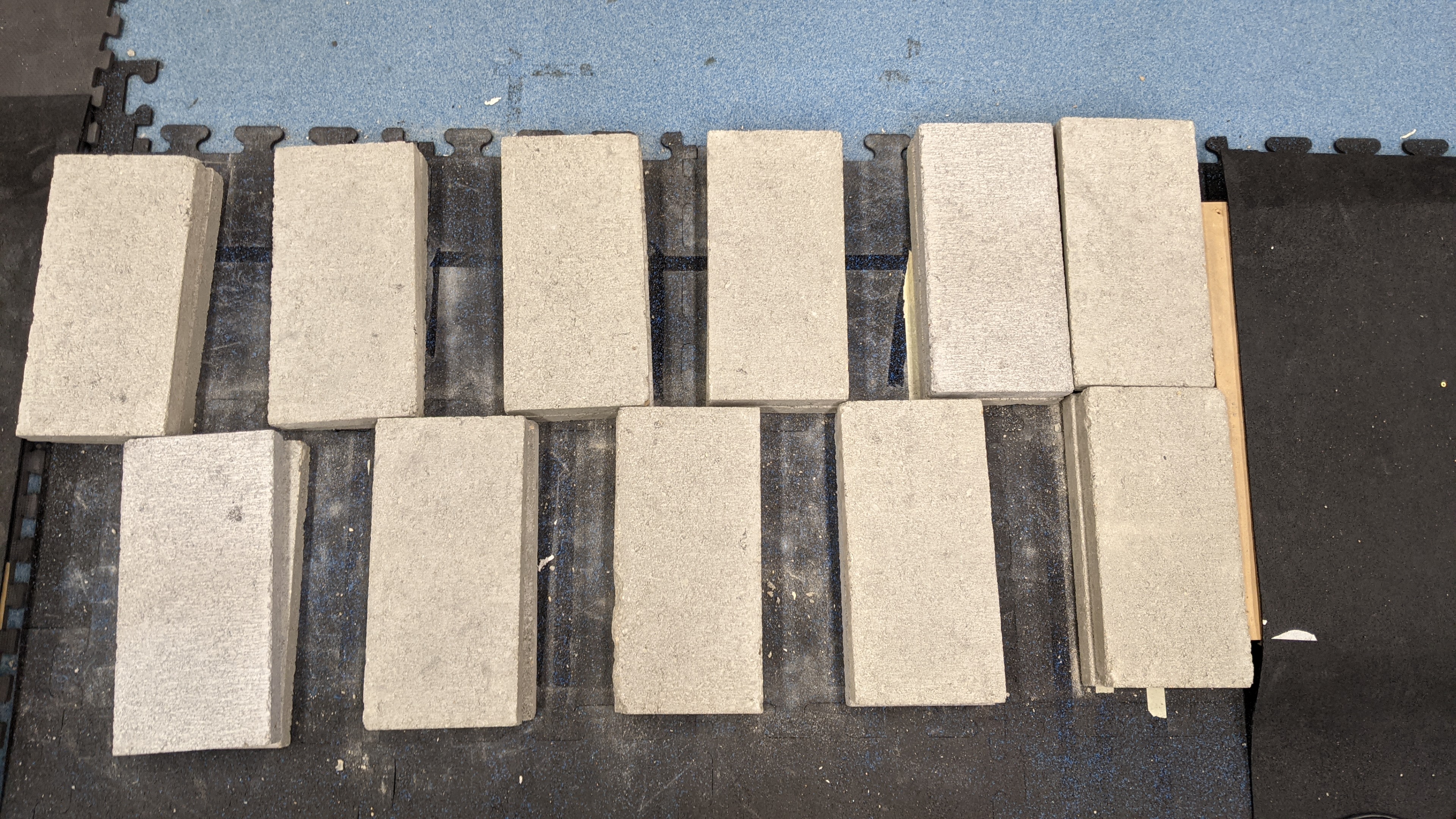}
        \caption{Staggered Terrain}
        \label{fig:staggered}
    \end{subfigure}
    \begin{subfigure}[b]{0.23\textwidth}
        \centering
        \includegraphics[width=0.95\textwidth]{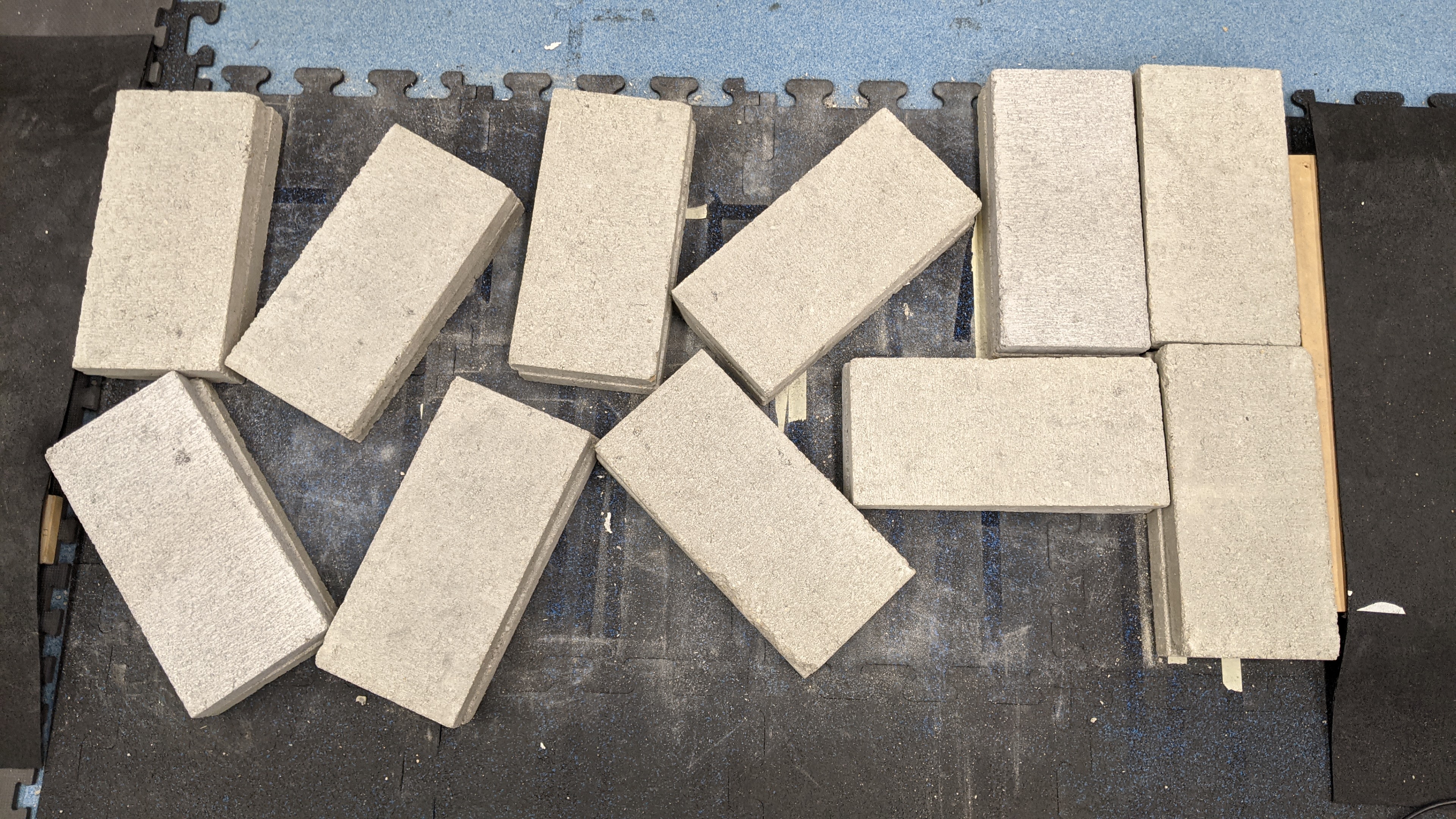}
        \caption{Random Terrain}
        \label{fig:random}
    \end{subfigure}
    \begin{subfigure}[b]{0.23\textwidth}
        \centering
        \includegraphics[width=0.95\textwidth]{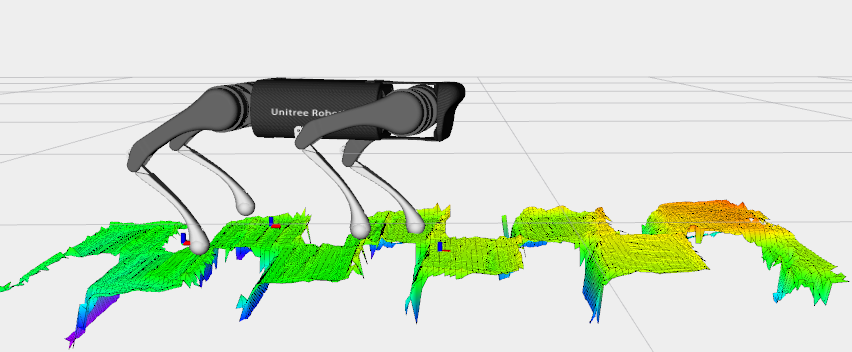}
        \caption{A1 robot on discrete terrain, visualized from real-world data.}
        \label{fig:robot}
    \end{subfigure}
    \caption{\small Different terrains tested in our experiments, and visualization of a local map built on the robot. 
    }
    \label{fig:experiment_environment}
\end{figure}
 The discrete-time equations of motion for the rigid-body dynamics are then, 
%  \vspace{-0.4cm}
  \begin{align}
    \position_{k+1} = \position_k +  \velocity_k \timestep, \label{eq:discrete_pos_dyn}\\
    \velocity_{k+1} = \velocity_k +\timestep \grav + \frac{\rblinforce_{k+1}}{\mass} \timestep, \label{eq:discrete_vel_dyn}\\
    \rotationmat_{k+1} = \rotationmat_k \deltarot_k,\label{eq:discrete_rotmat_dyn}\\
     \inertia \angularvel_{k+1} = \deltarot_k^T\inertia\angularvel_k + \timestep \rbmoment_{k+1}, \label{eq:discrete_angvel_dynamics}
 \end{align}
 where $\rotationmat_k \in SO(3)$ denotes the rotation matrix, $\inertia \in \mathbb{R}^{3 \times 3}$ is the inertia tensor, $\grav \in \mathbb{R}^3$ is the gravity vector. 
 $\deltarot_k \coloneqq \exp{\left( \timestep\hat{\angularvel}_k \right)}$ denotes the change in orientation of the body from time $t_k$ to time $t_{k+1}$, where the exponential map $\exp: \mathfrak{so}(3) \rightarrow SO(3)$ maps a skew-symmetric matrix to a rotation matrix.
 We define the state of the rigid body to be $\rbstate_k := \left[\position_k^T, \velocity_k^T, \rotationmat_k^T, \angularvel_k^T \right]^T$ and the input to be $\rbinput_k := \left[\rblinforce_k^T, \rbmoment_k^T \right]^T$.
 
 \textbf{{Linearization}}: Having obtained the discrete-time model of the system, we next compute a \emph{variation-based} linearization \cite{wu2015variation} of the nonlinear discrete-time dynamics around a reference trajectory. The resulting linearized model will be locally valid on the $SE(3)$ manifold, and will be used to formulate our MPC problem as a quadratic program (QP) that can be solved in real-time. To compute the linearization, we take \emph{infinitesimal variations} around a reference state.
 
 Since the position and velocity dynamics in \eqref{eq:discrete_pos_dyn} and \eqref{eq:discrete_vel_dyn} are already linear, we turn to the linearization of the orientation dynamics \eqref{eq:discrete_rotmat_dyn} and \eqref{eq:discrete_angvel_dynamics}. The variations on $SO(3)$ with respect to a reference trajectory $\rotationmat_k^d \in SO(3)$ is given by,
 \begin{equation}
     \var \rotationmat_k = \rotationmat_k^d \hat{\eta}_k, \label{eq:variation_rotmat}
 \end{equation}
\noindent
 where $\eta_k \in \mathbb{R}^3$ and $\hat{\eta}_k$ maps $\mathbb{R}^3 \rightarrow \mathfrak{so}(3)$ such that $\hat{a}b = a \times b$ for all $a, b \in \mathbb{R}^3$, where $\times$ is the vector cross
product. The variation in the angular velocity is 
\begin{equation}
    \var \angularvel_k = \frac{1}{\timestep}\left(\deltarot_k\eta_{k+1}-\eta_k\right). \label{eq:variation_angvel}
\end{equation}

Using the variations in \eqref{eq:variation_rotmat}, and from the nonlinear discrete-time dynamics of the rotation matrix in \eqref{eq:discrete_rotmat_dyn}, we get the linear discrete-time system about a reference as,
 \begin{align}
     \rotationmat_{k+1} &= \rotationmat_k \exp{\left(\hat{\angularvel}_k\timestep\right)},\\
     \var\rotationmat_{k+1} &= \var\rotationmat_k \exp{\left(\hat{\angularvel}_k^d\timestep\right)} + \rotationmat_k^d \var \exp{\left(\hat{\angularvel}_k \timestep\right)},\\
     \Rightarrow \eta_{k+1} &= \deltarot_k^{d^T}\eta_k + \timestep \deltarot^{d^T} \var \angularvel_k.
 \end{align}

Similarly, the linearized discrete-time dynamics for the angular velocity is obtained from \eqref{eq:discrete_angvel_dynamics}, \eqref{eq:variation_rotmat} and \eqref{eq:variation_angvel} as,
\begin{align}
\var \left(\inertia\angularvel_{k+1}\right) = \var \left(\deltarot_k^T \inertia \angularvel_k + \timestep \rbmoment \right),
\end{align}
\begin{align}
\inertia \var \angularvel_{k+1} = \var \deltarot_k^T\inertia \angularvel_k^d + \deltarot_k^{d^T}\inertia \var \angularvel_k + \timestep \var \rbmoment_{k+1},\\
\Rightarrow \inertia \var \angularvel_{k+1} = \deltarot_k^{d^T} \left(\timestep \reallywidehat{\inertia \angularvel_k^d }+ \inertia \right) \var \angularvel_k + \timestep \var \rbmoment_k.
\end{align}
Putting together the linear and angular components, the linearized discrete-time system is given by,
\begin{equation}
    \var \rbstate_{k+1} = A_k \var \rbstate_k + B_k \var \rbinput_k, \label{eq:linear_discrete_time_dyn}
\end{equation}
where $\var \rbstate_k \coloneqq \left[\var \position_k^T, \var \velocity_k^T, \eta_k^T, \var \angularvel_k^T \right]^T$ is the error state of the linearized system.
\begin{figure*}
    \centering
    \includegraphics[width=0.95\textwidth]{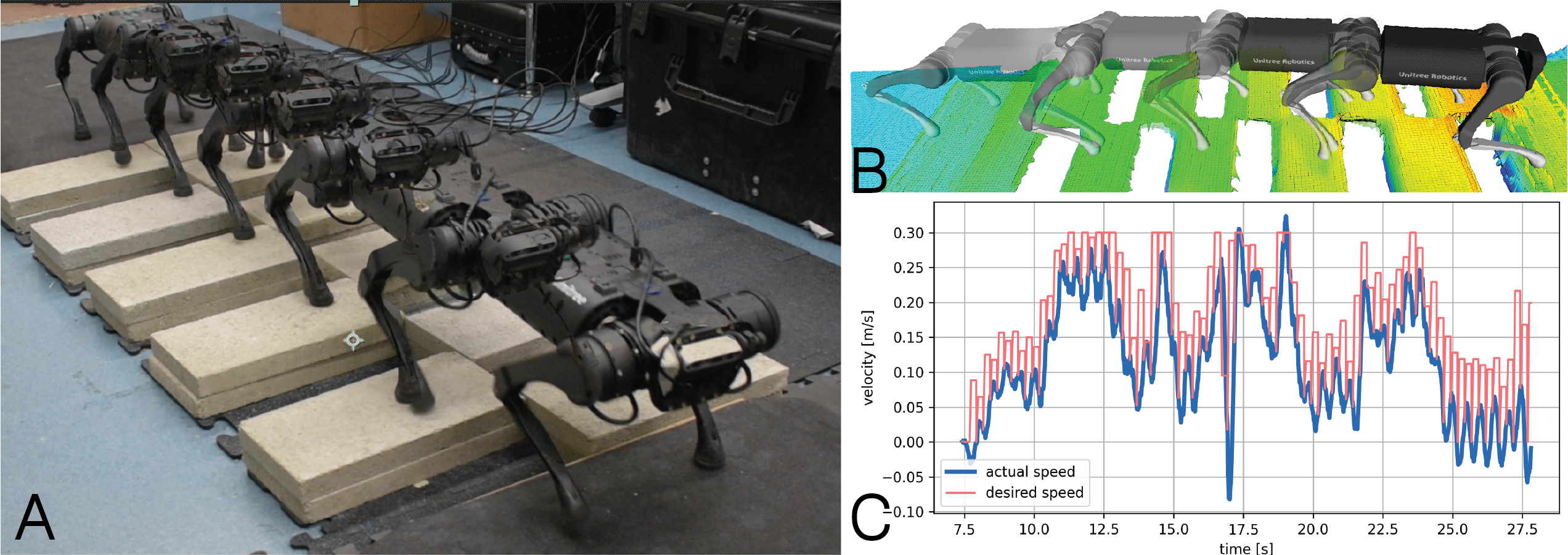}
    \caption{\small (A) Snapshots of the robot, (B) visualization of the terrain map illustrating the foot-placement of the robot on the stepping stones, and (C) forward velocity of the robot from real world data.}
    \label{fig:staggered_terrain}
\end{figure*}
The matrices $A_k$ and $B_k$ are given by,

\begin{align}
    A_k &\coloneqq \begin{bmatrix}
    \identity_3 & \timestep \identity_3 & \zeros_3 & \zeros_3\\
    \zeros_3 & \identity_3 & \zeros_3 & \zeros_3\\
    \zeros_3 & \zeros_3 & \deltarot_k^{d^T} & \timestep \deltarot_k^{d^T}\\
    \zeros_3 & \zeros_3 & \zeros_3 & a_{\angularvel}
    \end{bmatrix},\label{eq:angluar_vel_discrete}\\ 
    B_k \coloneqq &\begin{bmatrix}
    \zeros_3 & \zeros_3\\
    \frac{\timestep \identity}{m} & \zeros_3\\
    \zeros_3 & \zeros_3 \\
    \zeros_3 & \timestep \inertia^{-1}
    \end{bmatrix}, 
    a_{\angularvel} \coloneqq \inertia^{-1}\deltarot_k^{d^T}\left(\timestep\reallywidehat{\inertia \angularvel_k^d} + \inertia \right). \nonumber
\end{align}

The linear discrete-time dynamics in \eqref{eq:linear_discrete_time_dyn} represents the evolution of the infinitesimal variations on the manifold around a reference trajectory. These variations represent the distance between two points on the manifold. 
Under the assumption that the actual rotation matrix $\rotationmat_k$ is close to the desired rotation matrix $\rotationmat_k^d$, the variation $\var \rbstate_k$ can be approximated as 
\begin{align}
    \var \rbstate_k \approx  \begin{bmatrix}\position_k - \position_k^d \\
    \velocity_k - \velocity_k^d \\
    \frac{1}{2}\left(\rotationmat_k^{d^T}\rotationmat_k - \rotationmat_k^T\rotationmat_k^d\right)^\vee\\
    \angularvel_k - \rotationmat_k^T\rotationmat_k^d\angularvel_k^d \end{bmatrix}, \label{eq:error_approx}
\end{align}

\noindent where the vee map $\vee: \mathfrak{so}(3) \rightarrow \mathbb{R}^3$ is the inverse of the hat operator, so that $\hat{x}^\vee = x, ~ \forall x \in \mathbb{R}^3$. The last two terms in \eqref{eq:error_approx} denote the errors on the tangent bundle $TSO(3)$ manifold \cite{lee2010control,bullo2019geometric}. 
With this approximation, the dynamics in \eqref{eq:linear_discrete_time_dyn} represents the evolution of the error on the manifold locally around the reference trajectory $\rbstate^d$. 

\textbf{{Geometric MPC-QP}}: Given the desired CoM states $\xi_d$ generated from the motion library at the current trotting step, we compute the initial error state $\var\rbstate(0)$ as in \eqref{eq:error_approx} and solve the following QP, 

\begin{align}
    \quadinput* &= \argmin_{\quadinput,  \var \rbstate_k, \var\rbinput_k} \|\var \rbstate_N \|_P + \Sigma_{k=0}^N \left(\|\var \rbstate_k \|_Q + \|\var\rbinput_k\|_R \right) \nonumber\\ 
    \text{s.t.} \qquad &\var \rbstate_{k+1} = A_k(\rbstate^d_k) \var\rbstate_k + B_k(\rbstate^d_k) \var\rbinput_k, \\
    &\quadinput \in \frictioncone, \label{eq:friction_cone}\\ 
    & 0 \leq {\quadinput^z}_{_i} \leq c_i \bar{\lambda}, \quad i\in \lbrace 0, 1, 2, 3\rbrace
    \label{eq:unilateral_constraint}\\
    &\graspmap \quadinput =  \var \rbinput_0 + \begin{bmatrix}\mass\grav\\\zeros_{3\times1}\end{bmatrix}, \\ \label{eq:wrench_conversion}
    &\var\rbstate_0 = \var\rbstate(0),
\end{align}

\noindent
where \eqref{eq:friction_cone} denotes the linearized friction-cone constraint, \eqref{eq:wrench_conversion} denotes the CoM wrench and contact forces, with $\graspmap$ denoting the grasp-map \cite{murray2017mathematical}. \eqref{eq:unilateral_constraint} represents the unilateral constraints on the vertical ground reaction forces at the feet; $c_i \in \lbrace 0, 1 \rbrace$ denotes the binary contact state of foot $i$. The above QP outputs the desired contact forces $\quadinput^*$. For legs in swing, the contact forces are set to zero by the constraint in \eqref{eq:unilateral_constraint}. We implement the above QP using the \texttt{OSQP} solver \cite{osqp}, with a horizon length of 10 and time-step of 0.05s, which can be solved at $1e^{-4}s$. The stance-leg torques are obtained through the quasi-static relation $\tau_{st} = -J_c^T\quadinput^*$.
\vspace{-0.2cm}

%% file: results.tex
\section{Experiments}

We demonstrate the robustness of our approach on the \texttt{Unitree A1} quadruped (Fig. \ref{fig:robot}) on a diverse set of terrains with discrete footholds (Fig. \ref{fig:experiment_environment}). These terrains consist of concrete blocks of size $\inch{6}\times\inch{16}$. The gap lengths between blocks range between $7cm$ and $18cm$, and can be in different orientations. The robot is required to move forwards while avoiding the gaps. The gap lengths are the same for the left and right legs in the \emph{aligned} terrain (Fig \ref{fig:aligned}), different in \emph{staggered} terrain (Fig. \ref{fig:staggered}) or random in \emph{random} terrain (Fig. \ref{fig:random}). Random terrains pose additional constraints on the lateral foot placement. The nominal commanded velocity is $0.25m/s$ and is updated by the gait library based on the desired foot position.

First, we compare our approach to the baseline in \cite{xie2021glide} (\emph{Heuristic}) which uses the closest stepping location to a Raibert-like footstep and a Jacobian linearized rigid-body model, without any motion libraries. We use the implementation in \cite{motionimitation}. This baseline tests the robustness of our approach over other heuristic approaches from literature shown on discrete terrain walking. 
Next, we incorporate motion libraries to this baseline stance controller (\textit{Jacobian with Gait Library}) and query CoM velocity and footstep location from the motion library. This experiment illustrates the need for geometric MPC on uneven terrain. Together, these experiments study the performance of our whole framework, against heuristic approaches from literature, as well as the importance of geometric MPC on uneven terrain. Table \ref{tab:comparison} summarizes the success rates of the three controllers on different terrains, over 3 hardware runs on the A1 robot.
\begin{table}
\scriptsize
    \centering
    \begin{tabular}{l c c c} \toprule
           &  Heuristic & Jacobian with Gait library & GVMPC (ours) \\
         \midrule
         Aligned       & $0/3$  & $1/3$ &  $3/3$ \\
         Staggered      & $0/3$  & $0/3$ &  $2/3$ \\
         \bottomrule
    \end{tabular}
    \caption{\small Success rates of the three controllers on different terrains over 3 hardware runs on the A1 robot. Our approach (GVMPC) outperforms the baseline controllers on aligned and staggered terrains. The failure mode of GVMPC on the staggered terrain is due to the stance foot slipping at the edge of the terrain. All controllers use the same vision feedback.}
    \label{tab:comparison}
\end{table}

We observe that the Heuristic approach is not able to successfully navigate any of the terrains. This is because the robot needs to speed up or slow down depending on the size of the gaps. Since the Heuristic baseline only changes the footstep position but maintains a constant CoM velocity, it is easily destabilized when walking over large gaps. The second baseline which uses the gait library is able to cross the aligned terrain in 1 trial, but fails on the staggered terrain. The Jacobian linearized model does not regulate the CoM velocities and orientations well in our experiments, causing the robot to go unstable. The instability is caused more in the lateral direction pointing towards foot placement feedback going unstable due to lateral and roll angular velocities. The failure mode in the staggered experiment for the GVMPC is due to the stance foot slipping at the edge of the terrain. 

Additionally, we conduct two runs of experiments on the random terrain, which is significantly more complicated and needs precise foot placement, and CoM position and orientation planning. Our approach is able to navigate this terrain in $2/2$ experiments. These experiments demonstrate that our proposed Geometric MPC is able to robustly stabilize the robot from a larger set of states around a desired trajectory. 

%% file: discussion.tex
\vspace{-1em}
\section{Conclusions}

In this paper, we present a planning and controls framework for vision-aided navigation for quadrupedal robots in challenging terrain. The method leverages offline computation of library of gaits parametrized by step lengths, and an on-board geometric MPC that takes into account the underlying geometric structure of the reduced-order rigid-body model, in both the discretization and linearization of the dynamics. Combining our proposed method with existing state-of-the-art tools for localization and mapping, we demonstrate successful implementation of quadruped locomotion on discrete terrain. While the primary focus of this work is locomotion on discrete terrain with varying step lengths, our method can potentially be extend to terrains with varying step widths as well as for turning on discrete terrain. 

A drawback of our method is it requires the elevation map to be segmented into \emph{stepable} and \emph{un-stepable} regions, which is currently achieved by thresholding the height and normal vector direction of the elevation map and providing a safety margin from the edge of a stone. For small scale robots such as the A1, a gap between two stepping stones can be occluded in the resulting depth image due the low nominal height of the robot. This can result in inaccurate segmentation of the height map if the threshold and safety margins are not chosen appropriately. Our approach also utilizes an invariant EKF to estimate the position of the robot on the elevation map. Any drift in the position estimate between steps can lead to an inaccurate foot placement. The EKF on the A1 robot was particularly challenging to tune due to its compliant feet and the behavior of the contact sensor located at the foot on different surfaces. 